\documentclass[conference]{IEEEtran}
\IEEEoverridecommandlockouts
\usepackage{cite}
\usepackage{amsmath,amssymb,amsfonts}
\usepackage{algorithmic}
\usepackage{graphicx}
\usepackage{textcomp}
\usepackage{xcolor}
\def\BibTeX{{\rm B\kern-.05em{\sc i\kern-.025em b}\kern-.08em
    T\kern-.1667em\lower.7ex\hbox{E}\kern-.125emX}}
\begin{document}

\title{Reinforcement Learning for Cognitive Delay/Disruption Tolerant Network Node Management in an LEO-based Satellite Constellation\\
}

\author{\IEEEauthorblockN{1\textsuperscript{st} Xue Sun}
\IEEEauthorblockA{\textit{University of Chinese Academy of Sciences} \\
\textit{Key Laboratory of Space Utilization, Technology and Engineering Center for Space Utilization, Chinese Academy of Sciences}\\
Beijing, China \\
sunxue16@mails.ucas.edu.cn}
\and
\IEEEauthorblockN{2\textsuperscript{nd} Changhao Li}
\IEEEauthorblockA{\textit{University of Chinese Academy of Sciences} \\
\textit{Key Laboratory of Space Utilization, Technology and Engineering Center for Space Utilization, Chinese Academy of Sciences}\\
Beijing, China \\
lichanghao20@mails.ucas.ac.cn}
\and
\IEEEauthorblockN{3\textsuperscript{rd} Lei Yan}
\IEEEauthorblockA{\textit{Key Laboratory of Space Utilization, Technology and Engineering Center for Space Utilization, Chinese Academy of Sciences} \\
\textit{Technology and Engineering Center for Space Utilization, Chinese Academy of Sciences}\\
Beijing, China \\
yanlei@csu.ac.cn}
\and
\IEEEauthorblockN{4\textsuperscript{th} Suzhi Cao}
\IEEEauthorblockA{\textit{Key Laboratory of Space Utilization, Technology and Engineering Center for Space Utilization, Chinese Academy of Sciences} \\
\textit{Technology and Engineering Center for Space Utilization, Chinese Academy of Sciences}\\
Beijing, China \\
caosuzhi@csu.ac.cn}

}

\maketitle

\begin{abstract}
In recent years, with the large-scale deployment of space spacecraft entities and the increase of satellite onboard capabilities, delay/disruption tolerant network (DTN) emerged as a more robust communication protocol than TCP/IP in the case of excessive network dynamics. DTN node buffer management is still an active area of research, as the current implementation of the DTN core protocol still relies on the assumption that there is always enough memory available in different network nodes to store and forward bundles. In addition, the classical queuing theory does not apply to the dynamic management of DTN node buffers. Therefore, this paper proposes a centralized approach to automatically manage cognitive DTN nodes in low earth orbit (LEO) satellite constellation scenarios based on the advanced reinforcement learning (RL) strategy advantage actor-critic (A2C). The method aims to explore training a geosynchronous earth orbit intelligent agent to manage all DTN nodes in an LEO satellite constellation scenario. The goal of the A2C agent is to maximize delivery success rate and minimize network resource consumption cost while considering node memory utilization. The intelligent agent can dynamically adjust the radio data rate and perform drop operations based on bundle priority. In order to measure the effectiveness of applying A2C technology to DTN node management issues in LEO satellite constellation scenarios, this paper compares the trained intelligent agent strategy with the other two non-RL policies, including random and standard policies. Experiments show that the A2C strategy balances delivery success rate and cost, and provides the highest reward and the lowest node memory utilization.
\end{abstract}


\section{Introduction}
The future space network will not only involve various types of physical nodes, such as low/medium/geosynchronous earth orbit satellites, deep-space probes, planetary surface spacecraft, Etc., but also deal with mission challenges with different quality of service (QoS) requirements, such as command, telemetry, and scientific data applications. Recent advances in communications technology, energy efficiency, launch costs, and conveyor-belt manufacturing have allowed for enhanced onboard capabilities and more risky/innovative approaches to deployment\cite{r1}, so NASA expects to realize the architecture of cognitive network in the future space communication infrastructure\cite{r2}. Reinforcement learning (RL) is a method of making decisions in a trial-and-error manner that does not require a dataset, which rewards decisions that yield positive outcomes to learn the good and the bad in a given environment. It is widely used in the literature to solve the Markov decision process (MDP)\cite{r3}. However, the learning process of RL itself needs to explore the entire system to achieve the best policy, so it is unsuitable for large networks. In order to break through its limitations in application practice, deep neural networks have opened up a new era for the development of RL. Deep reinforcement learning can achieve complex network optimization solutions, allowing network entities to learn and build knowledge about the communication and network environment, providing autonomous decision-making, and significantly increasing learning speed, especially in problems with large state and action spaces\cite{r4}.

The delay/disruption tolerant network (DTN) architecture is very similar to the Internet architecture, except that a bundle layer is added between the transport layer and the application layer. The DTN protocol, similar to the Internet protocol (IP), is called the bundle protocol (BP). The standardization of the BP has been carried out for many years. DTN has been standardized and defined in IETF RFC5050, and the blue book standard has been released, as well as CCSDS standards 734-B-1 and 734.2-B-1. The BP in the DTN architecture abstracts the differences in the underlying protocols so that different protocols can be adapted. For example, the closer detector nodes on the planet's surface can use the standard TCP-based network, and the long-distance link between the planetary relay satellites can use the LTP. The BP layer is concerned with the custody transmission and routing of data and also has a rate control function to support congestion prediction and avoidance. In addition to the BP, DTN also defines an opportunistic discovery mechanism named IP neighbor discovery (IPND)\cite{r16}. The IPND mechanism allows previously unknown DTN nodes to exchange link information to allow them to start communicating. Thus, without knowledge of node addressing and scheduling, discovery beacons will be used to exchange information in UDP datagrams if two nodes come into contact with each other.

This paper attempts to solve the DTN routing buffer management problem in an LEO-based satellite constellation scenario through machine learning methods. The mechanism mentioned above characteristics of DTN lay the foundation for combining with RL methods under long-delay/disruption links in complex space-based unknown networks. Although the contact plan in the contact graph routing (CGR)\cite{r19} strategy can provide prior knowledge, which is designed by NASA JPL and has been able to prove sufficient accuracy and efficiency to become the fact routing framework for space DTN\cite{r20}, the impact of vast and diverse traffic/missions on DTN node management is unknown and unpredictable, such as buffer overflow or congestion. RL can provide optimal support for this problem scenario. The strategy proposed in this paper is based on an implementation of a cognitive delay/disruption tolerant node (CDTN)\cite{r5}, which enables RL agents and DTN nodes to interact and make decisions without human involvement. RL agents can sense node and network state changes to autonomously trigger specific actions to achieve management goals and be immune to expected failures/congestion, such as dropping bundles, changing node radio data rates, modifying routes path, or doing nothing.


The main contributions of this paper are as follows:

\begin{itemize}
\item Based on the DtnSim simulation platform and Python, an LEO-based satellite constellation scenario that conforms to real astrodynamics is created, which means that the physical nodes participating in the control of intelligent agents are larger and scalable.
\item The advanced RL method advantage actor-critic (A2C) is introduced into the scene, and a centralized intelligent agent model is designed to manage the DTN autonomously, which maximizes the delivery success rate and minimizes the network resource consumption cost while considering the node memory utilization. This intelligent agent not only dynamically adjusts the radio data rate but also performs drop operations based on data priority.
\item Validation of the above concepts and algorithms is implemented, and experimental analysis is performed to compare the performance of the A2C strategy with other non-RL policies.
\end{itemize}

\section{Background and Related Works}

\subsection{Node and Resource Management in CGR}

Routing is one of the most critical components of DTN architecture, and CGR is already a mature routing technology applied in the field of space-based DTN. In the standard CGR protocol, the bundles are classless and are served with a first-come-first-serve(FCFS) policy from the buffer by default, which is assumed to have infinite capacity\cite{r6}. However, in order to improve QoS and delivery success rate, the bundle marked as critical in the bundle information can be copied and sent to the destination node through multiple paths\cite{r18}, at the cost of increasing the transmission energy, contact capacity occupancy, and node memory utilization and other resources. Moreover, performance evaluations of DTN protocols using standard CGR are based on nodes having unlimited available memory to store and forward data\cite{r7}. Therefore, an RL formula for DTN node and resource management is needed to cope with the future dynamic changes in the network, data flow, and node failures. The motivation of this paper is to use an advanced RL method to fill the research gap that nodes in the cognitive DTN can autonomously manage node buffers to avoid overflow while improving QoS and reducing the number of resources consumed by the system.

\subsection{RL and A2C}

\begin{figure}[htbp]
\centerline{\includegraphics[width=0.9\linewidth]{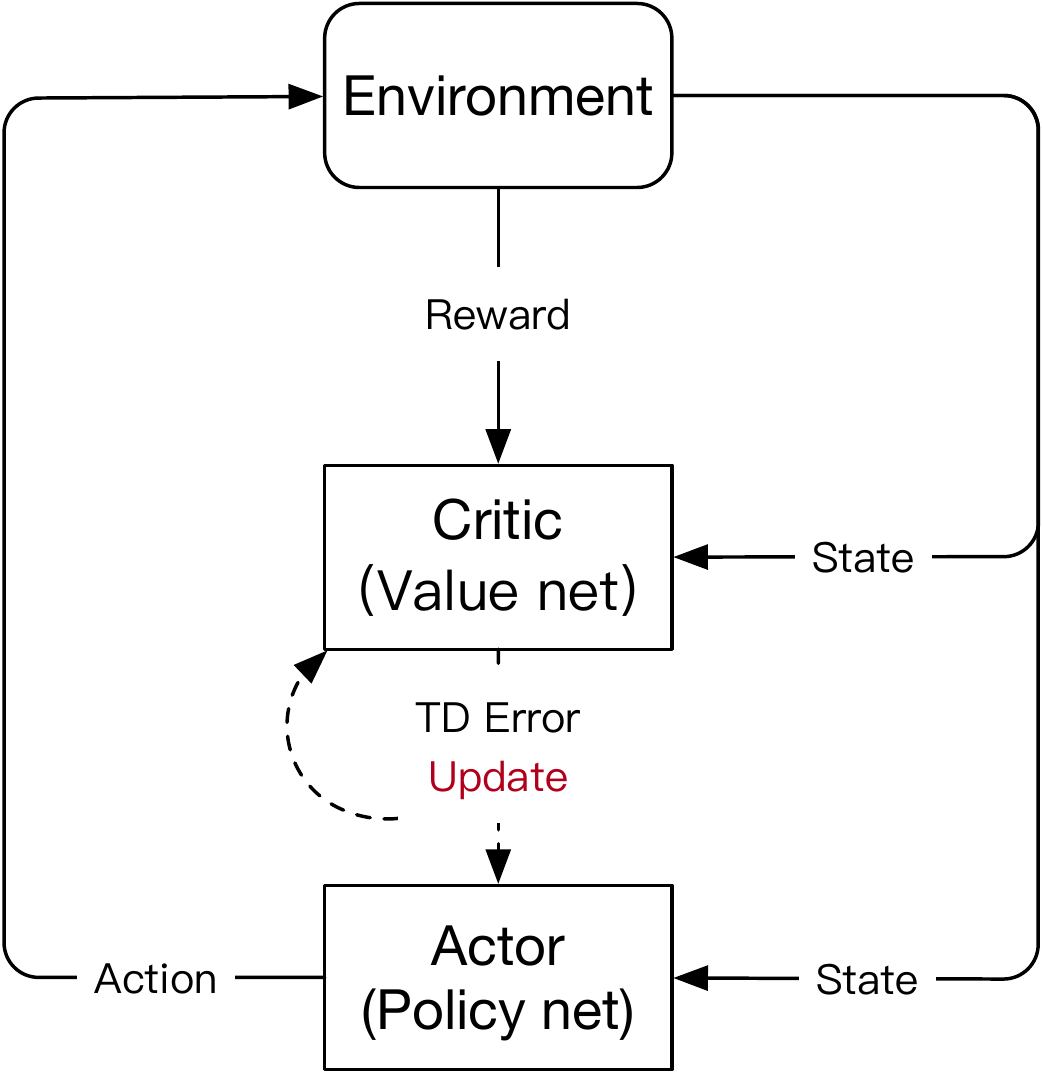}}
\caption{Schematic diagram of the principle of Actor-Critic algorithm.}
\label{Fig1}
\end{figure}

The core idea of RL agents is high-scoring orientation, learning the correspondence between data and labels and the relationship between the two through repeated attempts in the environment, and selecting high-scoring actions as much as possible. RL includes many algorithms, and this paper chooses a synchronous, deterministic variant of the asynchronous advantage actor-critic (A3C)\cite{r8} method proposed by Google DeepMind in 2016, namely A2C. The A2C method is a kind of actor-critic algorithm, and it combines the traditional policy gradients algorithm and the value-based Q-Learning algorithm. It can not only overcome the shortcomings of the former round update and reduce the learning efficiency but also can perform a single-step update and help pick the proper action in a sequence of actions that would be paralyzed by the latter. Actor and critic are two different systems that are trained at the same time and can be replaced by different neural networks. The former is used to predict behavior probability, and the latter is used to predict the value of state input. The critic sees the potential reward of the current state by learning the relationship between the environment and the reward and uses the TD error to guide the actor to update the action step by step, as shown in Fig.~\ref{Fig1}. 

\subsection{Applications of RL in DTN}
Over the past decade, RL has begun to be applied to network access and rate control, including communication and networking, caching and offloading, security and connectivity preservation, and miscellaneous problems such as traffic routing, resource scheduling, and data collection\cite{r4}. First, we focus on space-based DTN using deterministic routing techniques instead of the flood-type opportunistic routing approach commonly employed on terrestrial networks\cite{r9}. \cite{r10} has demonstrated that applying RL to complement the decision-making of CGR in DTN can outperform traditional shortest path algorithms in various complex environments. Of particular relevance to our work is the application of RL to resource allocation problems that cause congestion and degrade the QoS in systems with limited capacity. \cite{r11} introduced a congestion control algorithm based on Q-Learning and game theory in the scenario composed of terrestrial Internet of things and several satellites. However, the focus of this paper is on DTN node buffer control management, which is more complicated than the M/M/1/K queuing model, because it is necessary to consider that the receiving rate of the bundle is not constant, the traffic situation changes with time, and the contact time between nodes is not continuous, Etc. This paper is mainly based on the following essential works. Although \cite{r12} pays attention to the research gap of RL application in DTN node buffer management and considers dynamic traffic, the proposed system model is based on the fact that the contact between nodes in the network is constant, not considering node mobility and multi-priority traffic. \cite{r13} tried to apply the deep Q-learning method to the DTN node buffer management in the earth observation constellation scenario based on \cite{r12}, and added the action space category. The main difference between this paper and the above work is that this paper attempts to apply the advanced RL method A2C to the LEO-based satellite constellation scenario and extend the state space and reward function. The goal is to optimize the delivery success rate while entirely using contacts, links' capacity, and memory resources to solve the problem of resource green utilization and QoS guarantee.

\section{Problem Formulation}

\subsection{System Model}

The system model constructed in this paper is suitable for the LEO-based satellite constellation composed of Delta configuration, all LEO nodes are CDTN, and the geosynchronous earth orbit (GEO) satellite group completes the entity of the centralized intelligent agent. The task type supports three priorities (the RL agent can selectively drop certain packets when congested), all DTN entity nodes have a fixed upper limit on the maximum buffer size, the radio data rate is controllable,  and the contacts between nodes are under the celestial dynamics model considering the natural mobility of the nodes. Network factors such as memory status, data rate, contact/link status, Etc. of all LEO nodes are available to the GEO intelligent agent. Below, we describe the RL agent proposed in this paper by three essential aspects: state space, action space, and reward function.

\subsection{RL Agent}

\subsubsection{State Space}

For the state space definition of the problem formulation, it is assumed that the network parameters defined in the state space are always available, and these network parameters need to characterize the current state of the DTN nodes as successfully as possible, allowing the intelligent agent to act accordingly, with the ultimate goal of maximizing the expected cumulative reward. In addition to what is commonly thought of as node memory utilization, radio data rate, Etc., adding, e.g., contact/link occupancy and average delay of packets to the state space representation, is critical for intelligent agents to adapt to different traffic/task conditions dynamically.

Therefore, the always available network condition of the RL agent designed in this paper is characterized by the following vector:

\begin{equation}
s=[O_C,R_i,D_A,U_i], i=1,2,\dots,24
\end{equation}
Where $O_C$ is the occupancy rate of all contacts/links in the time step, which is equal to the cost described later divided by the total link capacity, $R_i$ is the radio data rate of DTN node i , $D_A$ is the average waiting delay of all successfully delivered packets in this time step, and $U_i$ is the memory utilization of DTN node i at the current time step.

\subsubsection{Action Space}

For the action space definition of the problem formulation, a centralized intelligent agent arranged in the GEO satellite group takes six different actions according to the current state of all LEO satellite DTN nodes in the network:

\begin{itemize}
\item \#1: Double the radio data rate of all DTN nodes at the same time ($\leq$500bit/s).
\item \#2: Halve the radio data rate of all DTN nodes at the same time.
\item \#3: Drop low priority packets of all DTN nodes at the same time.
\item \#4: Drop low and medium priority packets of all DTN nodes at the same time.
\item \#5: Drop all priority packets of all DTN nodes at the same time.
\item \#6: Do nothing.
\end{itemize}

\subsubsection{Reward Function}

The agent's goal is to maximize the delivery success rate of packets considering the multi-priority task scenario while minimizing the contact/link occupancy cost and controlling the memory utilization of all DTN nodes. Therefore, the calculation method of the reward function designed in this paper is as follows:

\begin{equation}
R(s,a)=f(\max_{i=1,\dots,24}U_i)\cdot \frac{Delivered_{bits}}{Cost_{bits}}
\end{equation}
Where $Delivered_{bits}$ is the number of bits successfully delivered to the destination, $Cost_{bits}$ is the number of bits allocated (i.e., cost) for all contacts/links, and $f(\max_{i=1,\dots,24}U_i)$ is a penalty factor calculated as follows:

\begin{equation}
f(\max_{i=1,\dots,24}U_i)=\frac{1}{1+exp(-a(b-max_{i=1,\dots,24}U_i)}
\end{equation}
Where a=25, b=0.3.

\section{Experimental Procedure}

\subsection{Scenario Case Study}

\begin{figure}[!t]
\centering
\includegraphics[width=0.9\linewidth]{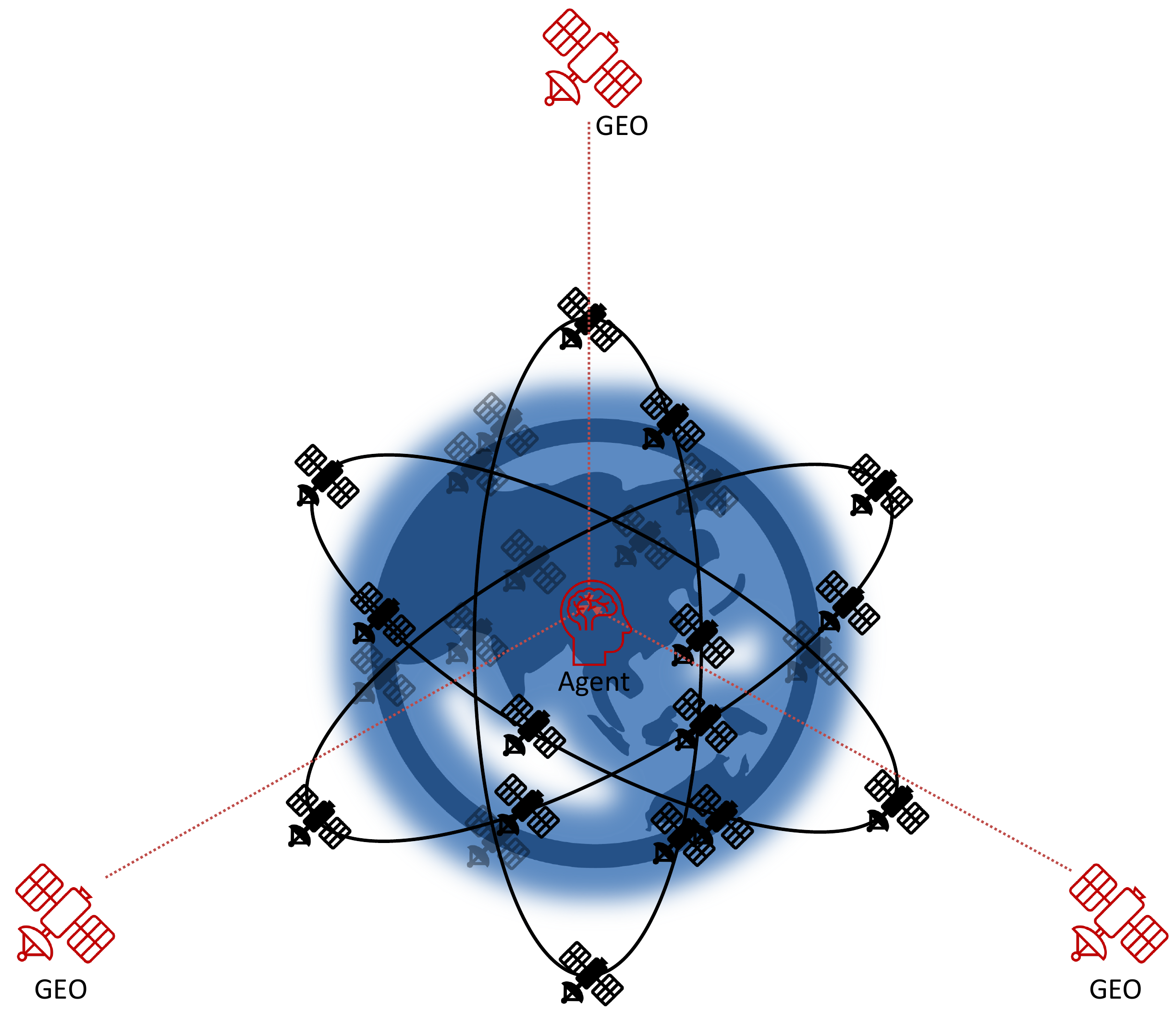}
\caption{Schematic diagram of a DTN with an LEO-based satellite constellation implementing the global communication service system as a training/test scenario, where the centralized intelligent agent is jointly formed by three GEO satellites, indicated by the red icon.}
\label{Fig2}
\end{figure}

In this paper, we aim to explore the performance of the A2C approach in scenarios targeting the realization of an LEO-based satellite constellation for the global communication services system. As shown in Fig.~\ref{Fig2}, the scenario contains 27 nodes, of which the configuration of 24 LEO satellites belongs to the Delta-type constellation. Each orbital plane has 8 LEO satellites and 3 orbital planes, the orbit height is 710km, and the orbit inclination angle is $98.5^{\circ}$. The other three GEO satellites use a centralized approach to train RL agents that can simultaneously control all LEO DTN satellite nodes.

\subsection{Simulation Environment: DtnSim}

We need to build an environment so the agent can get feedback from the environment and learn from it. For this paper's scenario, it is necessary to simulate the traffic transmission between different system nodes and accurately represent the parameters listed in the state space of all entities that make up the system. Therefore, this paper adopts the Python-based simulator for DTN named DtnSim\cite{r17} developed by JPL, which is currently an open-source package distributed under the Apache 2.0 license. The routing strategy implemented by DtnSim is CGR. We set the episode duration (training and evaluation) to 8000s (200 steps) and the maximum buffer size of satellite network nodes to 80Kbits. At the same time, the "AgcPacketGenerator" in DtnSim is used to create packets with priority. A single DTN node generates a packet every 9 seconds. Each packet has three priorities and a TTL of 3600s. From the global perspective of GEO satellites, the packet generation of the entire network is random.

\subsection{Training and Evaluation of the RL Agent}

We use the Python-based stable-baselines library \cite{r15} to train 1000 episodes. Each episode contains 200-time steps, and each time step is 40 seconds. The settings of an episode and step size refer to the fact that DTN experts believe it is reasonable to change the network parameters no less than every 30s\cite{r12}. All simulation parameters and training hyperparameters are listed in the Table.~\ref{Table1}. The discount factor is set to 0.99 because we are interested in high reward values in the future, and it is a commonly used value in the literature. The radio rate is continuously randomized at the beginning of each episode to improve training efficiency and enable the RL model to explore more network states as quickly as possible.

\begin{table}[htbp]
\caption{DtnSim simulation parameters and Stable-Baselines A2C training hyperparameters}
\begin{center}
\begin{tabular}{|ll|}
\hline
\multicolumn{2}{|c|}{DtnSim Parameters}                                      \\ \hline
\multicolumn{1}{|l|}{Episode duration(training and evaluation)} & 200 steps  \\
\multicolumn{1}{|l|}{Maximum buffer size of network nodes}      & 80K bits   \\
\multicolumn{1}{|l|}{Maximum radio data rate}         & 500 bits/s \\ \hline
\multicolumn{2}{|c|}{Stable-Baselines Parameters}                            \\ \hline
\multicolumn{1}{|l|}{Action time step}                          & 40 s        \\
\multicolumn{1}{|l|}{Discount factor($\gamma$)}                 & 0.99       \\
\multicolumn{1}{|l|}{Number of training episodes}               & 1000       \\
\multicolumn{1}{|l|}{Number of evaluation episodes}             & 100        \\
\multicolumn{1}{|l|}{Learning rate($\alpha$)}                   & $10^{-7}$  \\
\multicolumn{1}{|l|}{n\_steps}                                  & 5          \\ \hline
\end{tabular}
\label{Table1}
\end{center}
\end{table}

In order to evaluate the effectiveness of the intelligent agent using the A2C training strategy in the LEO-based satellite constellation scenario, we automatically save the strategy every five episodes during the training process and randomly select a strategy as the optimal strategy after the training converges, and evaluate the network performance for 100 additional episodes. Finally, the performance of this A2C strategy is compared with the performance of two non-RL benchmark policies:

\begin{itemize}
\item The standard policy, where the radio data rates of all satellite nodes are set to the maximum rate and do not change throughout the evaluation, represents the standard CGR strategy implemented by DtnSim.
\item The random policy takes random actions.
\end{itemize}

\section{Results}

\subsection{Convergence Performance and Operation Time of A2C Algorithm}

\begin{figure}[!t]
\centering
\includegraphics[width=0.9\linewidth]{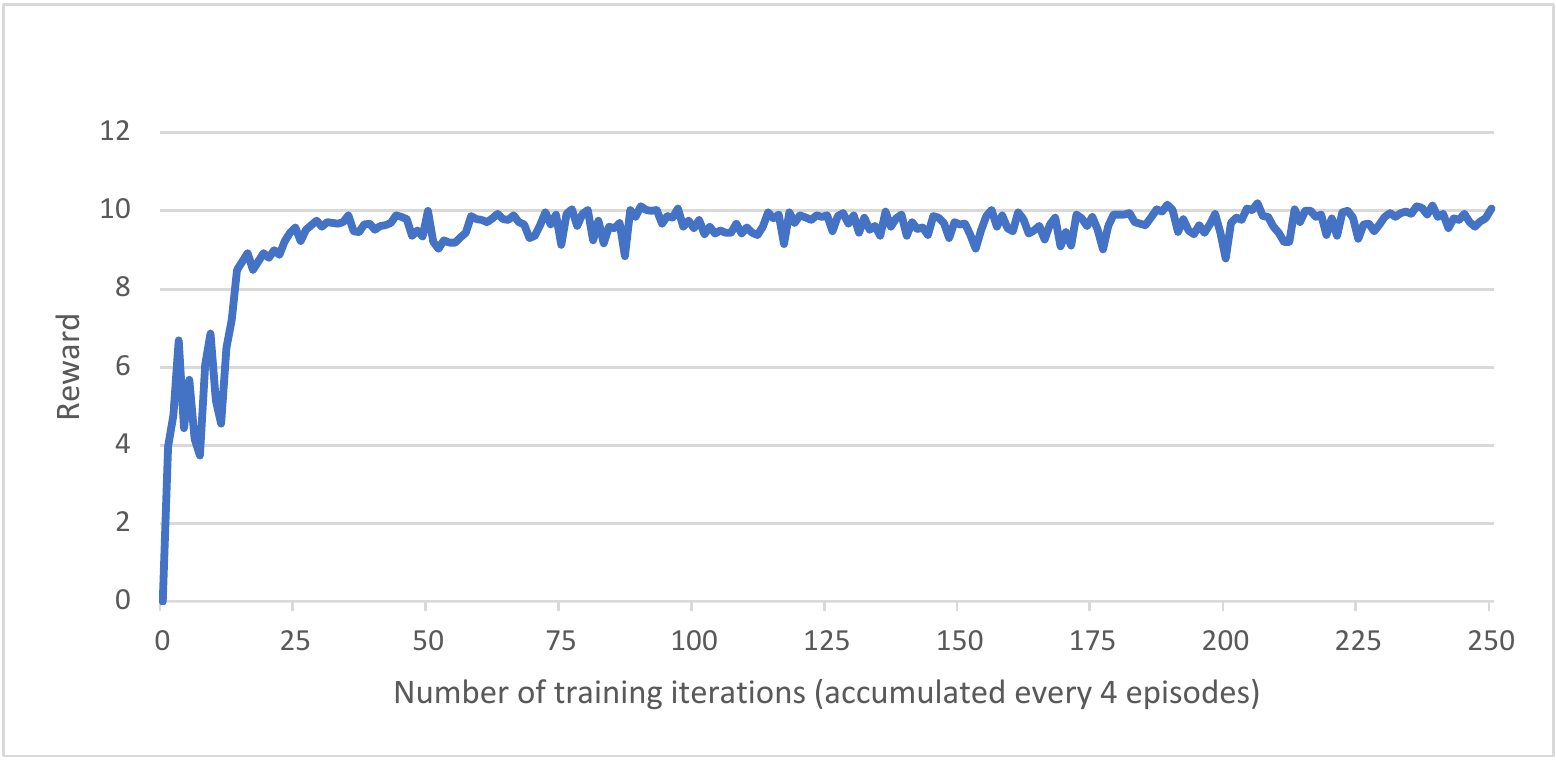}
\caption{Reward convergence of DTN intelligent agent training of 1000 episodes.}
\label{Fig3}
\end{figure}

RL training often uses the reward indicator to judge whether the RL algorithm has converged. When the accumulated reward within an episode does not change significantly, it can be regarded as training approximate convergence to ensure that the agent has approached the optimal policy. We use the accumulated reward every four episodes as the abscissa to show the convergence of the training process, as shown in Fig.~\ref{Fig3}. It can be observed that the DTN intelligent agent can find a strategy with a high reward within only 100 (25*4) episodes. Moreover, the following strategies have been stable at the reward value of about 10. Therefore, we can conclude that applying the A2C algorithm in this paper's scenario can converge to a good policy.

In practical application, the algorithm of RL will face slow operation speed, and real-time update operation is required when the environment changes. Therefore, the training process of this algorithm in the application can be performed synchronously on the terrestrial nodes, and the trained strategy can be duly added to the satellite node for the evaluation process. The implementation of this simulation relies on a MacBook Air notebook equipped with an Apple M1 chip with a memory of 16GB. Under the single-core computing setting, the training process for 1000 episodes takes 100 hours and 11 minutes, an average of 6 minutes per episode. During the evaluation, the A2C algorithm takes an average of 0.6 milliseconds to make corresponding actions in each step, which indicates that the algorithm model is designed to operate at a fast speed and has practical application prospects. The reason for the big difference between the operation time of the training process and the evaluation process can be explained that the environmental computing of the DtnSim simulation software used in the training process takes up too much time, and when the actual satellite node executes the direct test of the optimal strategy, no environment simulation is required. Therefore, separate statistics are made for the response time of a single step, which can characterize the actual satellite onboard operation conditions as closely as possible.

\subsection{Evaluation of the Memory Utilization, Data Rate and Dropped Bundle Size of All LEO Satellite Nodes under Different Strategies with Step Progress}

\begin{figure}[!t]
\centering
\begin{minipage}{1\linewidth}
        \centerline{\includegraphics[width=0.9\linewidth]{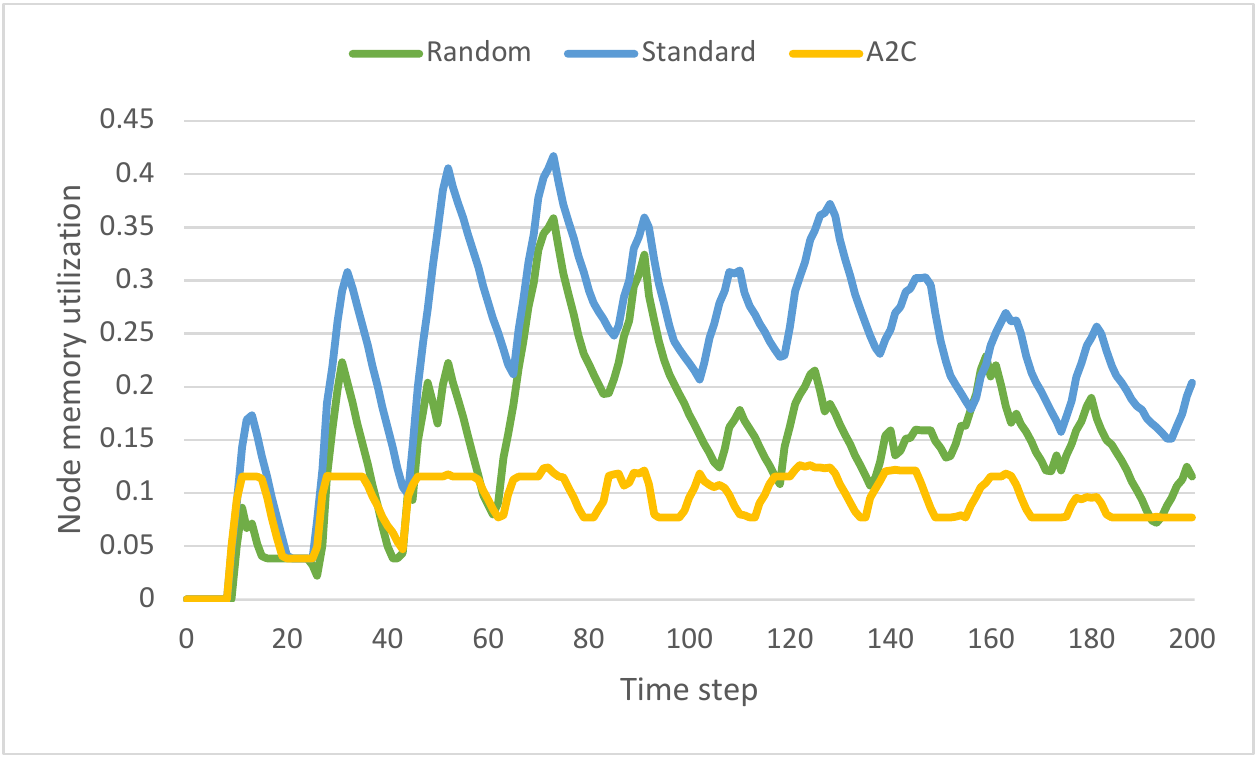}}
        \centerline{(a)}
\end{minipage}

\begin{minipage}{1\linewidth}
        \centerline{\includegraphics[width=0.9\linewidth]{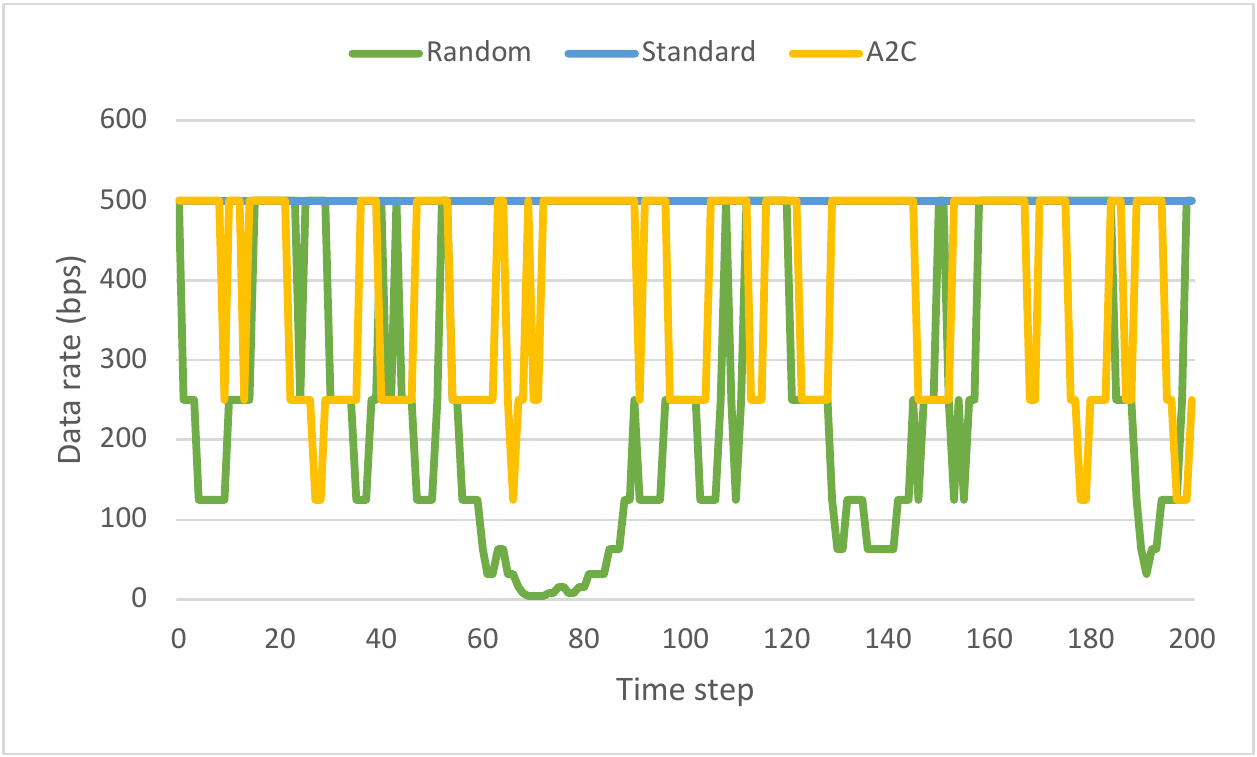}}
        \centerline{(b)}
\end{minipage}

\begin{minipage}{1\linewidth}
        \centerline{\includegraphics[width=0.9\linewidth]{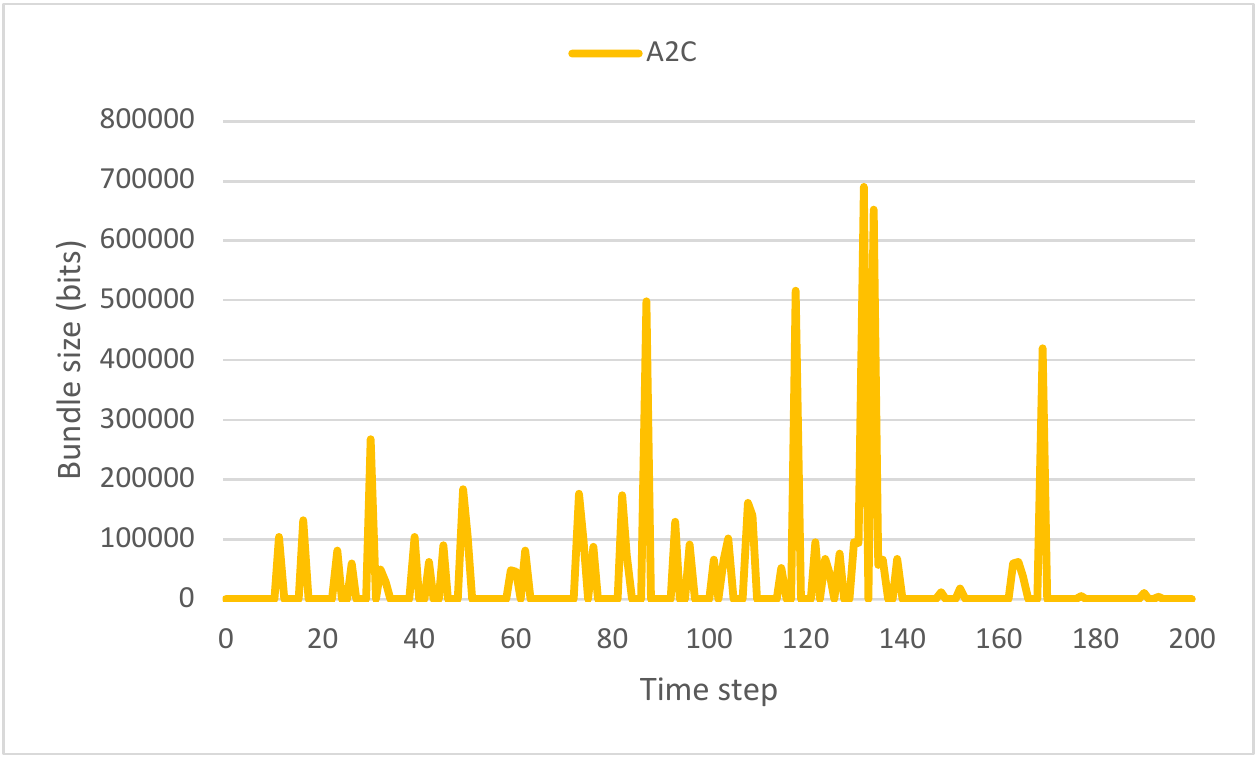}}
        \centerline{(c)}
\end{minipage}
\caption{Changes in (a) node memory utilization (max), (b) data rate (max) of all LEO satellites, and (c) bundle size dropped due to $\#3\sim 5$ actions within an episode.}
\label{Fig4}
\end{figure}

Fig.~\ref{Fig4} shows the network performance of the trained DTN agent and two other non-RL strategies as the step size progresses within an episode, including node memory utilization, data rates of all satellite radios, and the action of the intelligent agent at the same time. As can be seen from Fig.~\ref{Fig4}, the A2C strategy shows the best performance and maintains a stable and lowest memory utilization rate, and continuously adjusts the radio data rate or performs packet loss actions as the step size progresses. It is worth noting that the standard policy has always maintained the highest data rate (blue line), and here only the packet loss status artificially set in the action space is shown. Other automatic packet drop conditions in the system implemented by DtnSim have no difference between the three strategies, which is not specifically shown.

\subsection{Evaluation of the Performance of Reward, Cost, Delivery Success Rate, and Node Memory Utilization of A2C Intelligent Agent Compared with Two Non-RL Policies}

\begin{figure}[!t]
\centering
\begin{minipage}{1\linewidth}
        \centerline{\includegraphics[width=0.9\linewidth]{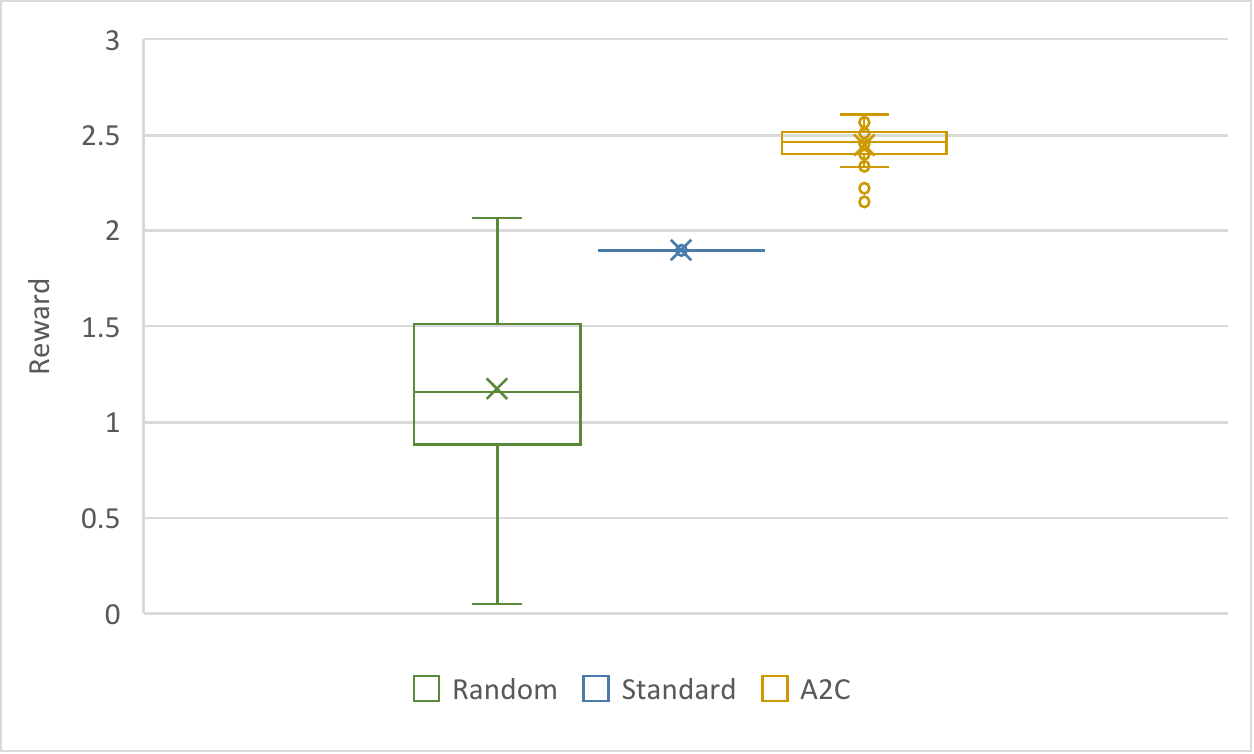}}
        \centerline{(a)}
\end{minipage}
\begin{minipage}{1\linewidth}
        \centerline{\includegraphics[width=0.9\linewidth]{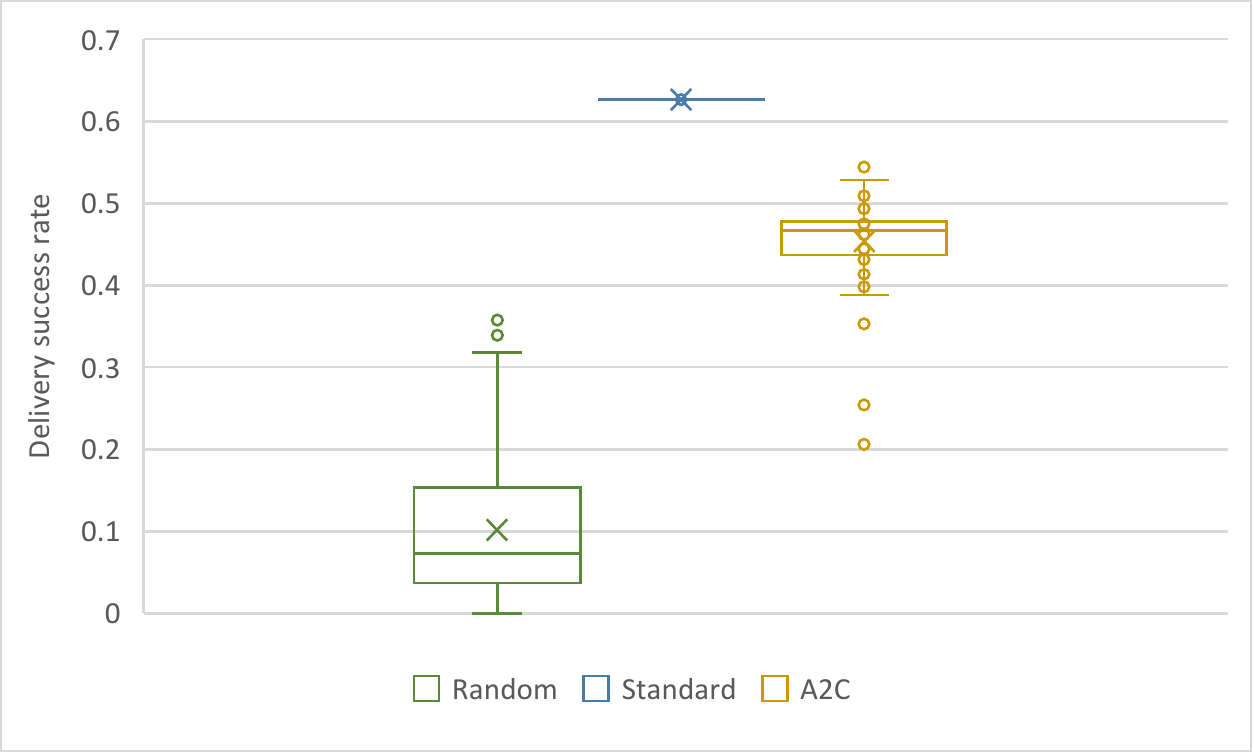}}
        \centerline{(b)}
\end{minipage}
\begin{minipage}{1\linewidth}
        \centerline{\includegraphics[width=0.9\linewidth]{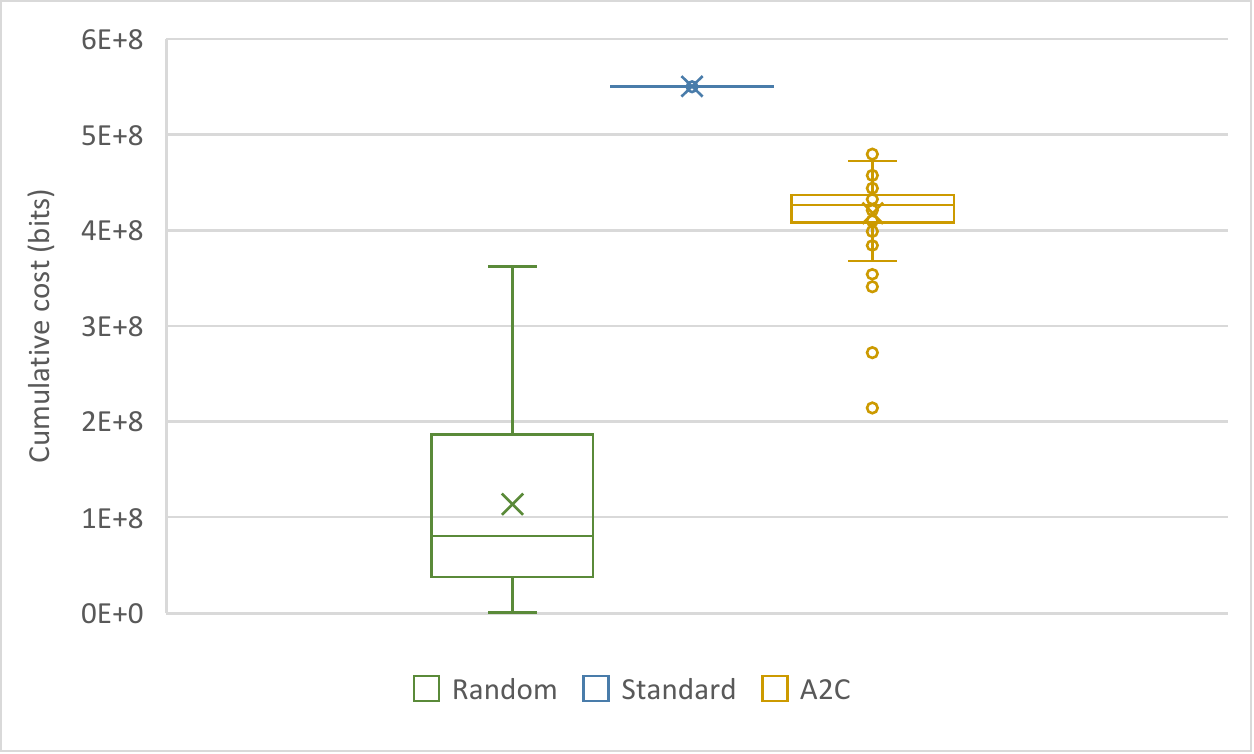}}
        \centerline{(c)}
\end{minipage}
\begin{minipage}{1\linewidth}
        \centerline{\includegraphics[width=0.9\linewidth]{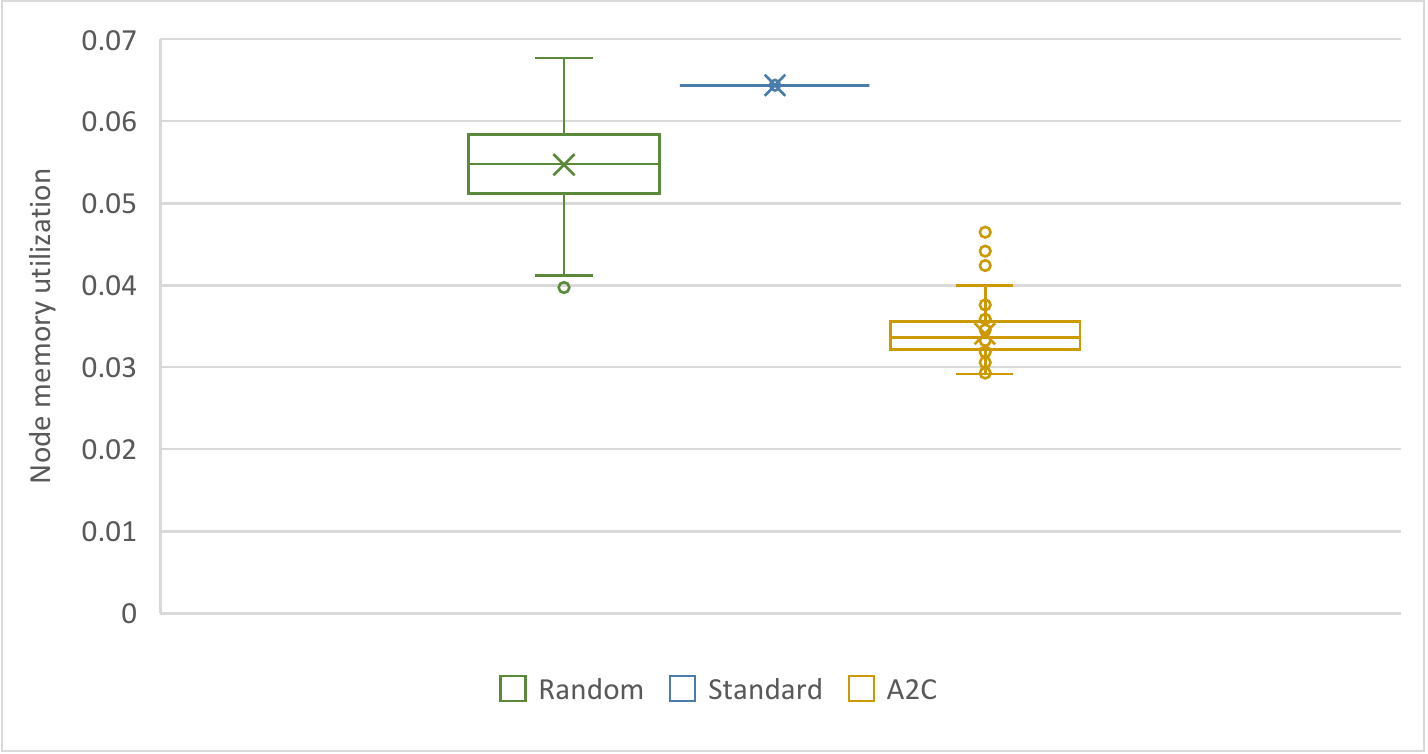}}
        \centerline{(d)}
\end{minipage}
\caption{Comparison between A2C and non-RL policies (Random and Standard) across (a) reward, (b) delivery success rate, (c) cumulative cost, and (d) node memory utilization for 100 evaluation episodes.}
\label{Fig5}
\end{figure}

Fig.~\ref{Fig5} compares the performance of the A2C intelligent agent, random and standard three strategies under 100 evaluation episodes, showing the cumulative reward, delivery success rate, cost, and node memory utilization rate of a single episode, respectively. The main conclusions that can be summarized from the figure are as follows:

\begin{itemize}
\item Compared with the other two non-RL policies, the A2C strategy provides the highest reward, maintains the lowest node memory utilization rate, and balances the relationship between the delivery success rate and cost, which is in line with the design of the reward function above.
\item The A2C strategy provides a higher reward and delivery success rate than the random strategy with lower node memory utilization.
\item Although the delivery success rate of the A2C strategy is slightly lower than that of the standard policy, it also reduces the cost and node memory utilization.
\item Compared to the random strategy, the variance of the A2C strategy is much smaller, while the standard strategy has the smallest variance because it fixes the initial data rate.
\end{itemize}

\section{Conclusion}

This paper proposes a centralized method to automatically manage cognitive DTN nodes in LEO-based satellite constellation scenarios based on the advanced RL strategy A2C. The experimental results obtained by the A2C method based on the environment provided by DtnSim demonstrate that A2C can rapidly converge within the first 100 training episodes and keep learning to a near-optimal policy in the last 900 training episodes. During the evaluation, the A2C algorithm takes an average of 0.6 milliseconds to make corresponding actions in each step, which indicates that the algorithm model is designed to operate at a fast speed and has practical application prospects. Compared with the other two non-RL policies, the A2C strategy shows the best performance in node memory utilization, has been relatively stable, and as the step size progresses, the radio data rate is continuously adjusted, or the packet drop action is performed. In addition, A2C through 100 evaluation episodes also shows the highest reward, the lowest node memory utilization, a higher successful delivery rate than the random strategy, and a lower cost than the standard strategy. Therefore, it is concluded that the A2C strategy finds a good balance between QoS and cost and considers the efficient utilization of nodes' resources.

Despite the promising results obtained in this first attempt to manage cognitive DTN nodes using A2C, more research is necessary to improve the RL agent and more thoroughly compare it against state-of-the-art strategies. So the next step would be implementing other RL algorithms with more advanced strategies, a richer and more diverse set of scenes to train RL agents, and enlarged action spaces. Because the intelligent agent in this paper is a centralized architecture to manage all LEO satellites uniformly, it is a pity that the distributed asynchronous training of multi-agent cannot be realized\cite{r14}. Finally, it may be an exciting and promising research topic to apply other artificial intelligence technology to more comprehensive space scenarios in the future.

\bibliographystyle{IEEEtran}
\bibliography{IEEEabrv}

\end{document}